\def\year{2022}\relax
\documentclass[letterpaper]{article} 
\usepackage{aaai22}  
\usepackage{times}  
\usepackage{helvet}  
\usepackage{courier}  
\usepackage[hyphens]{url}  
\usepackage{graphicx} 
\usepackage{tikz}
\usepackage{natbib}  
\usepackage{caption} 
\DeclareCaptionStyle{ruled}{labelfont=normalfont,labelsep=colon,strut=off} 
\usepackage{amsmath}
\usepackage[english]{babel}
\usepackage{amsthm}
\usepackage{amssymb}
\theoremstyle{definition}
\newtheorem{definition}{Definition}
\usepackage{multicol}

\usepackage{algorithm}
\usepackage{algorithmic}

\usepackage{newfloat}
\usepackage{listings}
\floatstyle{ruled}
\newfloat{listing}{tb}{lst}{}
\floatname{listing}{Listing}
\title{NeuralArTS: Structuring Neural Architecture Search with Type Theory}

\author{
    Robert Wu\textsuperscript{\rm 1},
    Nayan Saxena \textsuperscript{},
    Rohan Jain  \textsuperscript{} 
}
\affiliations{
    \textsuperscript{}University of Toronto\\

     \textsuperscript{} ML Collective\\
    \textsuperscript{\rm 1}rupert@cs.toronto.edu 
}

\usepackage{scalerel}
\usepackage{bibentry}

\begin{document}

\maketitle 

\begin{abstract} Neural Architecture Search (NAS) algorithms automate the task of finding optimal deep learning architectures given an initial search space of possible operations. Developing these search spaces is usually a manual affair with pre-optimized search spaces being more efficient, rather than searching from scratch. In this paper we present a new framework called Neural Architecture Type System (NeuralArTS) that categorizes the infinite set of network operations in a structured type system. We further demonstrate how NeuralArTS can be applied to convolutional layers and propose several future directions.
\end{abstract}

\section{Introduction}

 Neural Architecture Search (NAS) has proven to be a complex but important area of deep learning research. The aim of NAS is to automatically design architectures for neural networks. Most NAS frameworks involve sampling operations from a search space \citep{zoph2016neural}.  For example, Efficient Neural Architecture Search (ENAS) uses a controller based on reinforcement learning (RL) to sample child networks \citep{pham2018efficient}. \citet{yu2019evaluating} and other recent works have identified flaws in NAS algorithms, motivating improved methods for network construction. Search spaces are usually manually developed pre-search, rely heavily on researchers' domain knowledge, and often involve trial and error; it's more of an art than a science. Additionally, the domain of network operations is infinite given the multitude of basic operations and hyperparameters therein. It's hard to know which operations produce better performance in learning tasks. In this paper, we introduce a framework to potentially improve search spaces using generation and heuristics.
 



\section{Neural Architecture Type System}

Artificial neural networks can be interpreted as a programming domain, where operations can be categorized into type systems. A type system $\mathcal T$ is a formal system in which every element has a \textit{type} $\tau$, which defines its meaning and the operations that may be performed on it \citep{coquandtype}. One intuitive property of networks is the shape of the data as it moves through layers. Classes of operations such as pooling or convolution layers have mappings between input/output (I/O) dimensions, which can range from totally flexible (can be placed anywhere in a network) to complex (perhaps requiring a specific input and output shape). Therefore, it is possible to categorize operations into a type system based on these I/O mappings, opening new possibilities for search space optimisation in NAS. 
\noindent This idea can be extended to network subgraphs, since they can be abstracted as a block of operations with compound dimension functions. A consequence is that layers and subgraphs are also interoperable.

\begin{figure}[ht]
    \centering
    \input{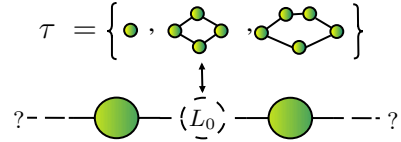}
    \caption{Possible interchange of $L_{\scaleto{0}{4.00pt}} \in \tau$ with other $\tilde L \in \tau
    $ or subgraph $[\tilde L_{\scaleto{1}{4.00pt}}, \ldots, \tilde L_{\scaleto{n}{4.00pt}}] \in \tau$. The unspecified topologies on either end represent previous/subsequent layers.}
    \label{fig:type_graphic}
\end{figure}

This idea is intuitive and not entirely new; it's explored in \citet{elsken2021bag}, albeit informally. To be formal and precise, the domain $\mathcal U$ of network operations can be categorized with a type system $\mathcal T$ centred around data shape compatibility. Compatibility is fundamental in informing which operations can precede, follow, or replace each other. For each operation layer $L$ with shape dimensions $\{1, \ldots, d\}$, define $I_{\scaleto{L}{4.00pt}} = (I_{\scaleto{L}{4.00pt}}^{(1)}, \ldots, I_{\scaleto{L}{4.00pt}}^{(d)})$ and $O_{\scaleto{L}{4.00pt}} = (O_{\scaleto{L}{4.00pt}}^{(1)}, \ldots, O_{\scaleto{L}{4.00pt}}^{(d)})$ to be the shapes of I/O data. Assume I/O shapes have the same number of dimensions for simplicity.

\subsection{Dimension Functions}

While these shapes can be constants, they are generally mappings that can be defined as a \textit{dimension function} $f_{\scaleto{L}{4.00pt}}$. Operation layer $L$ can have several other properties such as depth, stride, or bias values. Some or all of these properties may influence $O_{\scaleto{L}{4.00pt}}$, and they can be encapsulated in $f_L$. A dimension function can therefore be abstracted as $O_{\scaleto{L}{4.00pt}} := f_{\scaleto{L}{4.00pt}}(I_{\scaleto{L}{4.00pt}})$.

\subsection{Equivalence Properties}

NeuralArTS can centre around replacement/interchange of operation layers. Let $L_{\scaleto{A}{4.00pt}}, L_{\scaleto{B}{4.00pt}} \in \mathcal U$ be arbitrary layers with I/O dimensions $(I_{L_A}, O_{L_A})$ and $(I_{L_B}, O_{L_B})$ respectively.

\begin{definition}[Complete-Equivalence] 
$L_{\scaleto{A}{4.00pt}}$ and $L_{\scaleto{B}{4.00pt}}$ are \textit{completely equivalent} if all of their properties are equivalent. 
$L_{\scaleto{A}{4.00pt}} = L_{\scaleto{B}{4.00pt}} \iff (I_{L_A}, \ldots, O_{L_A}) = (I_{L_A}, \ldots, O_{L_B})$. 
\end{definition}

\begin{definition}[Type-Equivalence]
$L_A$ and $L_B$ are considered \textit{type-equivalent} if their I/O dimension functions are equivalent. In other words, $L_{\scaleto{A}{4.00pt}}$ and $L_{\scaleto{B}{4.00pt}}$ belong to the same type, $\tau$. $L_{\scaleto{A}{4.00pt}} \sim L_{\scaleto{B}{4.00pt}} \iff f_{L_A} = f_{L_B} \quad $
\end{definition} 

\begin{definition}[Instant-Equivalence]
$L_{\scaleto{A}{4.00pt}}$ and $L_{\scaleto{B}{4.00pt}}$ are \textit{instant-equivalent} at input size $I$ if their I/O dimension functions intersect at $I$. 
$L_{\scaleto{A}{4.00pt}} \perp_I L_{\scaleto{B}{4.00pt}} \iff f_{L_A}(I) = f_{L_B}(I) $
\end{definition}

\subsection{Sequential Compatibility}

Let $\mathcal I_{\scaleto{L}{4.00pt}}$ be the domain of acceptable input shapes into $L$, and $\mathcal O_{\scaleto{L}{4.00pt}}$ be the range of output shapes produced from $L$. 

\begin{definition}[Forward-Compatibility]
$L_{\scaleto{A}{4.00pt}}$ is \textit{forward compatible} to $L_{\scaleto{B}{4.00pt}}$ if all output shapes of $L_{\scaleto{A}{4.00pt}}$ are acceptable as input to $L_{\scaleto{B}{4.00pt}}$. $L_{\scaleto{A}{4.00pt}} \rightarrow L_{\scaleto{B}{4.00pt}} \iff \mathcal O_{L_A} \subseteq \mathcal I_{L_B}$
\end{definition}

\begin{definition}[Complete-Compatibility]
$L_{\scaleto{A}{4.00pt}}$ is \textit{completely compatible} to $L_{\scaleto{B}{4.00pt}}$ if they're mutually forward-compatible. $L_{\scaleto{A}{4.00pt}} \leftrightarrow L_{\scaleto{B}{4.00pt}} \iff L_{\scaleto{A}{4.00pt}} \rightarrow L_{\scaleto{B}{4.00pt}} \land L_{\scaleto{B}{4.00pt}} \rightarrow L_{\scaleto{A}{4.00pt}}$
\end{definition}

These properties regarding sequential compatibility of $L_{\scaleto{A}{4.00pt}}$ and $L_{\scaleto{B}{4.00pt}}$ are potentially useful in network construction; the controller in ENAS can change to consider compatibility in making direct and skip connections \citep{pham2018efficient}.

\subsection{Generative Example: Convolutional Layers}

\begin{figure}[h]
    \centering
    \input{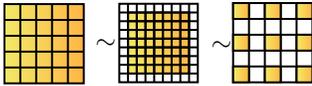}
    \caption{Type-equivalent convolutions that can be interchanged: a 5x5; a 7x7 with $p=1$; and a 3x3 with $d=2$.}
    \label{fig:conv_graphic}
\end{figure}

Let $\mathcal S_0$ be the original ENAS search space \citep{pham2018efficient} and $C \in \mathcal S_{\scaleto{0}{4.00pt}}$ be a convolution with kernel size $k$, padding $p$, and dilation $d$. These properties dictate type-equivalent convolutions. To test the efficacy of NeuralArTS, we first added some type-equivalent dilated variants of convolutions to $\mathcal S_0$. We found that adding even a single such convolution can outperform the baseline.
\begin{table}[htbp]
  \centering
 
  \resizebox{\columnwidth}{!}
  {
  \begin{tabular}{|c|c|c|}
    \hline
    \textsc{Search Space} & \textsc{Val Accuracy} & \textsc{Test Accuracy} \\
    \hline
    $S_0 \textnormal{ (Baseline)}$ & $80.47\%$ & $74.67\%$ \\
    \hline
    $S_0 + \textnormal{Conv}(k=3,p=2,d=2)$ & $\textbf{81.91\%}$ & $\textbf{78.64\%}$ \\
    \hline
    \end{tabular}}
    \caption{Performance of ENAS with a dilated convolution.}
    \label{tbl:dilation}
\end{table}
\begin{equation}
    f_{\scaleto{C}{4.00pt}}^{(i)}(I_{\scaleto{C}{4.00pt}}^{(i)}) := \left\lfloor\frac{I_{\scaleto{C}{4.00pt}}^{(i)} + 2p_{\scaleto{C}{4.00pt}}^{(i)} - d_{\scaleto{C}{4.00pt}}^{(i)}(k_{\scaleto{C}{4.00pt}}^{(i)} - 1) - 1}{s_{\scaleto{C}{4.00pt}}^{(i)}}\right\rfloor + 1
     \label{fig:io_conv_df}
\end{equation}

We introduce a generation technique that, without loss of generality, bounds two of $(k',p',d')$ to $(K,P,D)$ and derives the third using the convolution's I/O dimension function $f_{\scaleto{C}{4.00pt}}^{(i)}(I_{\scaleto{C}{4.00pt}}^{(i)})$ detailed in equation \ref{fig:io_conv_df}. Let $(k,p,d)$ be the properties of the original seed operation $L_0 \in \tau$. Let $(K,P,D)$ be the generation parameters, which include exactly one \texttt{None} and two positive integer ranges (inclusive). Algorithm \ref{alg:brute-force} explores the Cartesian product of $(K,P,D)$ to produce candidate tuples $(k',p',d')$ of properties. Each tuple's \texttt{None} value is replaced and derived from the other two values and the original properties $(k,p,d)$.

\begin{algorithm}[ht]
\footnotesize
\caption{\footnotesize GenerateTypeEquivalentConvs}
\label{alg:brute-force}
\begin{algorithmic}[1] 
\STATE \textbf{let} $settings = []$
\STATE \textbf{assert} $(K,P,D).\textnormal{count}(\texttt{None}) = 1$
\FOR {$(k',p',d') \in (K \times P \times D)$}
    \STATE \textbf{if} $K = \texttt{None}$ \textbf{then} \textbf{derive} $k' = \frac{2p' - 2p - d(k-1)}{d'+1}$
    \STATE \textbf{else if} $P = \texttt{None}$ \textbf{derive} $p' = \frac{d'(k'-1)}{2}$
    \STATE \textbf{else if} $D = \texttt{None}$ \textbf{derive} $d' = \frac{2p'}{k'-1}$
    \STATE \textbf{if} $k',p',d' \in \mathbb{N}$ \textbf{then} $settings$.append($(k',p',d')$)
\ENDFOR
\STATE \textbf{return} settings
\end{algorithmic}
\end{algorithm}

\section{Conclusion and Directions for Future Work}

This generation method can be improved with ``smarter" domains of operation properties; if performance proves to be continuous with respect to these properties, linear or manifold optimization might help generate more refined search spaces to speed up NAS. Another exciting prospect is that NeuralArTS can act as a heuristic for pre-optimized search spaces. It can naïvely eliminate completely-equivalent (or even type-equivalent) operations in preprocessing. More practical is changing the controller to modulate operation likelihoods at the type (rather than operation) level. If performance for types could be generalized, NeuralArTS can also be used to hierarchize NAS by performing shallow type-searches first, and then choosing random or ``best" representative(s) from each type. We hypothesize these directions might lead to improvements of NAS algorithms.

\section{Acknowledgements}
We would like to thank George-Alexandru Adam (Vector Institute; University of Toronto) \& Chuan-Yung Tsai (Vector Institute) for their comments and discussions that greatly influenced this paper. We are also grateful to Qi Jia Gao (University of Toronto) for assistance in the experimental setup. Finally, we thank the ML Collective community for the generous computational support, as well as helpful discussions, ideas, and feedback on experiments.

\bibliography{manuscript}
\newpage

\twocolumn[
  \begin{@twocolumnfalse}
    \begin{center}
        \LARGE{\textbf{Supplementary Material}}
        \bigskip
    \end{center}
  \end{@twocolumnfalse}
]


\section{Motivation: Dilated Convolutions}

Upon inspection of the Efficient Neural Architecture Search (ENAS) codebase, we discovered that the pre-optimized search spaces included only operations that had the same mapping of input-to-output data shapes \citep{pham2018efficient}. Intuitively, this pre-optimized search space $S_0$ was designed such that operations could be arranged and connected in any network structure and still remain compatible from input data to classification.


Our experiments entailed injecting type-equivalent convolutions to the original ENAS convolutions that had kernel size $k=3,5$. To be precise, we increased the padding $p$ and dilation $d$ values in these variant convolutions. Some experiments also include some dilated transposed convolutions. The experiments are labelled in this document as 3a/b/c/d/E/F/G for legacy reasons, and their specifications are outlined in Table \ref{tbl:dilation}.

\subsection{Outperforming Baseline in ENAS}

\begin{figure}[!t]
\centering
\includegraphics[width=.9\columnwidth]{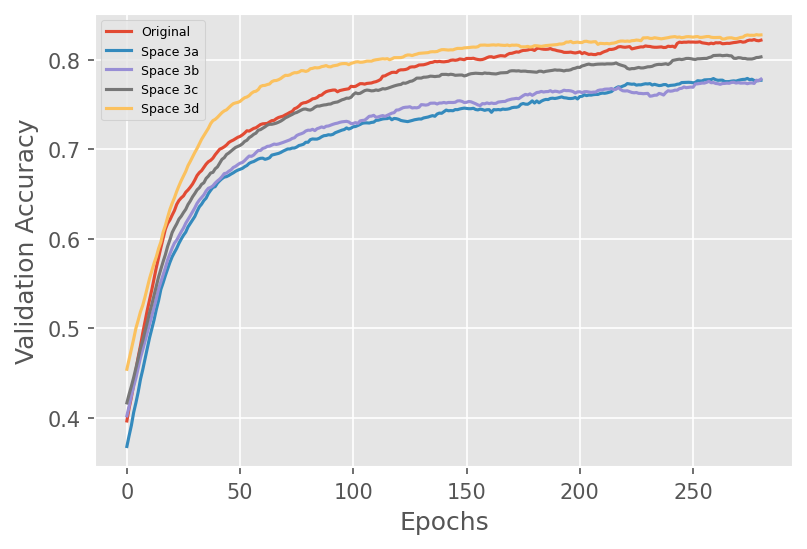}
\includegraphics[width=.9\columnwidth]{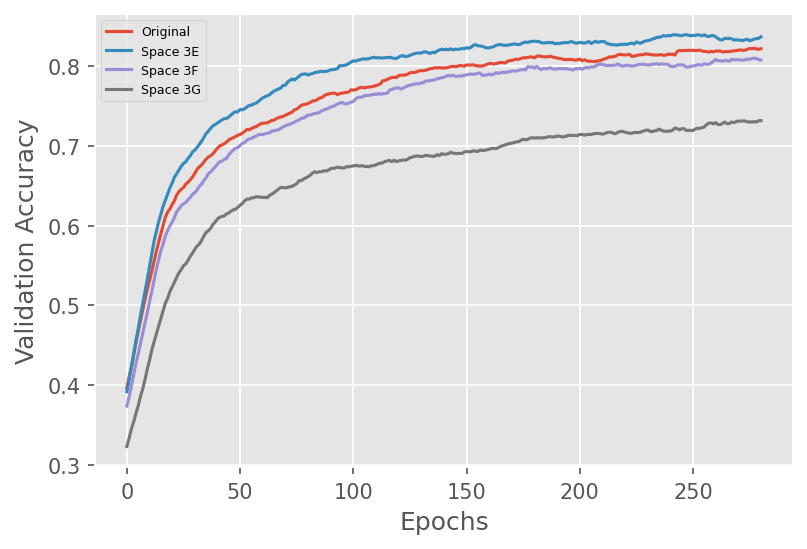}
\caption{Validation accuracy graphed over time of ENAS on various search spaces with added dilated convolutions. ENAS baseline is labelled Original in with a red graph. Note 3d and 3E outperform baseline in validation accuracy.}
\label{graphs}
\end{figure}

In fact, some of experiments were close to or outperformed the ENAS baseline. Results are listed in Table \ref{tbl:dilation} and validation accuracy over time is illustrated in Figure \ref{graphs}. Experiments 3d and 3E outperform the baseline in validation accuracy. In final test accuracy, 3c, 3d, and 3E outperform baseline by a few percentage points; 3E is able to achieve $78.64\%$ as compared to baseline's $74.67\%$.

\begin{table}[htbp]
\centering
\resizebox{\columnwidth}{!}{
    \begin{tabular}{|c|c|c|c|}
    \hline
    Code & \textsc{Search Space} & \textsc{Val Accuracy} & \textsc{Test Accuracy} \\ \hline
    Baseline & $S_0$ & $80.47\%$ & $74.67\%$ \\ \hline
    3a & $S_0 + \{\textnormal{Conv}(k=3,p=2,d=2),$ & &\\
    & $\textnormal{Conv}(k=3,p=3,d=3),$ && \\
    & $\textnormal{Conv}(k=5,p=4,d=2),$  & $73.24\%$ & $68.93\%$  \\
    & $\textnormal{Conv}(k=5,p=6,d=3),$ && \\
    & $\textnormal{Conv}(k=5,p=12,d=6)\}$ && \\
    \hline
    3b & $S_0 + 6\{\textnormal{Conv}(k=3,p=2,d=2)$ & & \\
    & $\textnormal{Conv}(k=3,p=3,d=3),$ && \\
    & $\textnormal{Conv}(k=5,p=4,d=2),$ &$74.26\%$ & $73.71\%$ \\
    & $\textnormal{Conv}(k=5,p=6,d=3),$ && \\
    & $\textnormal{Conv}(k=5,p=12,d=6)\}$ && \\
    \hline
    3c & $S_0 + 20\{\textnormal{Conv}(k=3,p=2,d=2)$ & & \\
    & $\textnormal{Conv}(k=3,p=3,d=3),$ && \\
    & $\textnormal{Conv}(k=5,p=4,d=2),$ &$78.53\%$ & $\textbf{76.64\%}$ \\
    & $\textnormal{Conv}(k=5,p=6,d=3),$ && \\
    & $\textnormal{Conv}(k=5,p=12,d=6)\}$ && \\
    \hline
    3d & $S_0 + 50\{\textnormal{Conv}(k=3,p=2,d=2)$ & & \\
    & $\textnormal{Conv}(k=3,p=3,d=3),$ && \\
    & $\textnormal{Conv}(k=5,p=4,d=2),$ & $\mathbf{80.49\%}$ & $\textbf{77.86\%}$ \\
    & $\textnormal{Conv}(k=5,p=6,d=3),$ && \\
    & $\textnormal{Conv}(k=5,p=12,d=6)\}$ && \\
    \hline
    3E & $S_0 + \{\textnormal{Conv}(k=3,p=2,d=2)\}$ & $\textbf{81.91\%}$ & $\textbf{78.64\%}$ \\
    \hline
    3F & $S_0 + \{\textnormal{Conv}(k=3,p=2,d=2)$ & $76.40\%$ & $70.26\%$ \\
    & $\textnormal{Conv}(k=3,p=3,d=3)\}$ && \\
    \hline
    3G & $S_0 + \{\textnormal{Conv}(k=3,p=2,d=2)$ & &\\
    & $\textnormal{Conv}(k=3,p=3,d=3),$ && \\
    & $\textnormal{Conv}(k=5,p=4,d=2),$ && \\
    & $\textnormal{Conv}(k=5,p=6,d=3),$ && \\
    & $\textnormal{Conv}(k=5,p=12,d=6),$ & $70.35\%$ & $68.11\%$  \\
    & $\textnormal{Conv}^T(k=3,p=2,d=2),$ && \\
    & $\textnormal{Conv}^T(k=3,p=3,d=3),$ && \\
    & $\textnormal{Conv}^T(k=5,p=4,d=2),$ && \\
    & $\textnormal{Conv}^T(k=5,p=6,d=3),$ && \\
    & $\textnormal{Conv}^T(k=5,p=12,d=6)\}$ && \\
    \hline
    \end{tabular}
}
\caption{Performance numbers of ENAS on various search spaces with added dilated convolutions. Bolded values in 3c, 3d, 3E outperformed baseline. Results are averaged over three runs.} 
\label{tbl:dilation}
\end{table}

\subsection{Further Inspiration}

It is not surprising that pre-optimized search spaces \citep{elsken2021bag} -- and the baseline of ENAS in particular \citep{pham2018efficient} -- are not actually optimal. It would be nearly impossible to create a search space that is the very best for the task because there likely exist infinite better search spaces not considered.

However, we didn't expect that simply adding a few naïvely dilated convolutions (especially in the case of 3E) would improve upon algorithms such as ENAS. The fact that these were easily found means that the problem of finding better search spaces might be tractable given a type system like we proposed.

\section{Search Space Generation}

One simple application of a type system is extending search spaces with generated operations. Such operations would be type-equivalent to existing operations in the original search space, which would ensure compatibility. This would automate the trial-and-error process we previously used in adding dilated convolutions.

\subsection{Brute-Force Search Space Generation}

As discussed in the main paper, the GenerateTypeEquivalentConvs algorithm bounds any two properties of $(k,p,d)$, and while exploring the Cartesian product of their ranges, the third is derived.

\begin{table}[htbp]
\centering
\resizebox{0.6\columnwidth}{!}{
    \begin{tabular}{|c|c|c|c|}
    \hline
    Code & $K$ & $P$ & $D$ \\ \hline
    \texttt{T1\_kd04} & $[1, 4]$ & \texttt{None} & $[1, 4]$ \\
    \texttt{T1\_kd08} & $[1, 8]$ & \texttt{None} & $[1, 8]$ \\
    \texttt{T1\_kd12} & $[1, 12]$ & \texttt{None} & $[1, 12]$ \\
    \texttt{T1\_kd16} & $[1, 16]$ & \texttt{None} & $[1, 16]$ \\
    \hline
    \texttt{T1\_kp04} & $[2, 4]$ & $[1, 4]$ & \texttt{None} \\
    \texttt{T1\_kp08} & $[2, 8]$ & $[1, 8]$ & \texttt{None} \\
    \texttt{T1\_kp12} & $[2, 12]$ & $[1, 12]$ & \texttt{None} \\
    \texttt{T1\_kp16} & $[2, 16]$ & $[1, 16]$ & \texttt{None} \\
    \hline
    \texttt{T1\_pd04} & \texttt{None} & $[1, 4]$ & $[1, 4]$ \\
    \texttt{T1\_pd08} & \texttt{None} & $[1, 8]$ & $[1, 8]$ \\
    \texttt{T1\_pd12} & \texttt{None} & $[1, 12]$ & $[1, 12]$ \\
    \texttt{T1\_pd16} & \texttt{None} & $[1, 16]$ & $[1, 16]$ \\
    \hline
    \end{tabular}
}
\caption{Generation parameters for brute-force search with bound starting from $1$ or $2$ ranging up to $4$, $8$, $12$, $16$.}
\label{tbl:brute-force}
\end{table}

\subsection{Our Files}

\begin{itemize}
    \item The files in which these extended search spaces can be found are in \texttt{enas\_types/}.
    \item The properties of the generate convolutions for each search space can be found in the folder \texttt{generated/}.
    \item Initial experimental results can be found in \texttt{ENAS-Experiments/results}.
\end{itemize}

\subsection{Experimental Setup}

The software used includes Python (3.6.x-3.8.x) and PyTorch (1.9), with CUDA (10.2, 11.1). Our hardware varies: CPUs included 2nd gen Xeon E, 3rd gen Core i5, 8th gen Core i3; GPUs included Nvidia GTX \{1050, 1050 Ti, 1080\}, RTX 2060, Tesla \{K80, P100 (Google Colaboratory)\}, TITAN Xp. Each experiment was processed with only one GPU with 3GB of allocated VRAM, and took between 18-36 real-world hours.

\end{document}